\documentclass[runningheads]{llncs}
\usepackage{graphicx}
\usepackage{amsmath,amssymb}
\usepackage{booktabs}
\usepackage{algorithm}
\usepackage{algorithmic}
\usepackage{comment}
\usepackage{multirow}
\usepackage{marvosym}

\begin{document}

\title{MedExpMem: Adapting Experience Memory for Differential Diagnosis}


\author{Qianhan Feng\inst{1}$^{*}$  \index{Feng, Qianhan}  \and
Zhongzhen Huang\inst{2}$^{*}$ \index{Huang, Zhongzhen} \and
Yakun Zhu\inst{2,3} \and
Yannian Gu\inst{2} \and
\\Winnie Chiu Wing CHU\inst{1}  \and
Xiaofan Zhang\inst{2,3}$^{\textrm{\Letter}}$  \and
Qi Dou\inst{1}$^{\textrm{\Letter}}$ }
\authorrunning{Q. Feng et al., submission to MICCAI 2026 review}
%
\institute{The Chinese University of Hong Kong\and
Shanghai Jiao Tong University \and
Shanghai Innovation Institue\\
\email{qianhan.feng@link.cuhk.edu.hk} \email{qidou@cuhk.edu.hk}\\
$^{*}$Equal contribution. $^{\textrm{\Letter}}$Corresponding authors.}
  
\maketitle

\begin{abstract}
Experienced physicians develop diagnostic expertise through clinical practice, acquiring not only disease knowledge but also the ability to differentiate confusable conditions. Current medical vision-language models (VLMs) lack this capability---their parameters encode static knowledge that does not evolve across diagnostic encounters. We propose \textbf{MedExpMem}, an experience memory framework enabling VLM-based diagnostic agents to accumulate differential diagnosis expertise. Unlike retrieval-augmented generation, which retrieves encyclopedic disease descriptions, MedExpMem memorizes discriminative experience derived from the agent's own diagnostic failures and organizes them as pairwise differential notes encoding key discriminators, actionable decision rules and reasoning error patterns. The framework adopts a two-phase construction process mirroring physician learning: initial practice exposes knowledge gaps, and reflective re-diagnosis refines understanding. When encountering new cases, the agent retrieves experience memory to guide differential reasoning. We evaluate MedExpMem on a radiology benchmark spanning 11 subspecialties. Results demonstrate consistent accuracy improvements, maximum 7.0\%, across diverse models and scales. Analytical experiments validate experience quality and robustness, demonstrating MedExpMem as a competitive method addresses medical adaptation needs beyond the reach of parameteric learning.

\end{abstract}

\section{Introduction}
\label{sec:introduction}

Medical expertise is not acquired solely through textbook knowledge; it is refined through repeated clinical practice. Although trainees may master canonical disease descriptions---typical imaging patterns, diagnostic criteria, and management guidelines---they often struggle to match the performance of experienced clinicians. The gap lies in clinical experience: discriminative, case-derived insights accumulated through repeated exposure to diagnostic ambiguity~\cite{schmidt1990cognitive}. Unlike textbook knowledge, which describes diseases in isolation, clinical experience encodes comparative judgments---why two conditions are easily confused, which subtle features distinguish them, and when one diagnosis should be favored. Such experience is not a verbatim archive of past cases but an abstraction distilled from prior diagnostic successes and errors.

Vision-language models (VLMs) have demonstrated strong capabilities in medical image interpretation~\cite{singhal2023large,zhang2023large}, yet operate in a fundamentally stateless manner: each diagnostic session proceeds independently without access to prior encounters. Models do not accumulate persistent experience from ongoing deployment. While fine-tuning could in principle encode experiential knowledge into parameters, it remains impractical in clinical settings due to privacy constraints, computational costs, and catastrophic forgetting risks---particularly for locally deployed hospital models requiring continuous adaptation.

Existing approaches address related aspects but do not enable persistent accumulation of differential experience. Retrieval-augmented generation (RAG) supplements models with external knowledge~\cite{lewis2020retrieval}, yet standard retrieval favors encyclopedic disease descriptions over the discriminative knowledge essential for differential diagnosis. Memory mechanisms for LLM-based agents~\cite{park2023generative,zhong2024memorybank,maharana2024evaluating,zhang2024memoryagent,zhao2024expel} typically store conversation histories or factual summaries rather than abstracting comparative reasoning patterns. Self-correction frameworks including Self-RAG~\cite{asai2023selfrag} and Reflexion~\cite{shinn2023reflexion} enable within-session improvement but lack persistent, cross-case memory. Neither paradigm addresses what medical 
diagnosis specifically demands: structured pairwise experience encoding why 
conditions are confused and how to distinguish them.

To bridge this gap, we propose \textbf{MedExpMem}, an experience memory framework enabling VLM-based diagnostic agents to accumulate differential expertise without parameter updates. Rather than organizing memory around isolated diseases, MedExpMem encodes knowledge at the level of diagnosis pairs---the fundamental unit of differential reasoning. Each pairwise experience note is derived from the agent's own diagnostic failures, capturing experience tailored to its specific blind spots. Notes include \textbf{key discriminators}, \textbf{actionable decision rules}, and \textbf{reasoning error patterns} that led to misdiagnosis. Critically, MedExpMem differs from RAG not only in what it retrieves, but in how memory is constructed: extracted from self-failures rather than external corpora.

Experience memory construction comprises two phases, both guided by 
expert-verified diagnostic knowledge. In \textbf{Phase~I}, the agent 
performs zero-shot diagnosis to expose its intrinsic reasoning blind spots. 
In \textbf{Phase~II}, the agent revisits erroneous cases with experience 
accumulated from Phase~I, simulating iterative practice to consolidate 
reliable patterns, correct inconsistent reasoning, and filter spurious 
errors. At inference, relevant experience is retrieved to help the agent 
recognize differential pitfalls it tends to overlook. This process mirrors 
clinical learning: initial exposure reveals weaknesses, deliberate practice 
refines judgment, and accumulated experience persists across cases. By 
decoupling experience from parameters, MedExpMem enables continual adaptation 
in privacy-sensitive medical environments.

We construct a benchmark from Eurorad~\cite{eurorad}, an educational radiology repository providing case data and expert guidance. This education-oriented setting closely mirrors experience accumulation scenarios and poses greater diagnostic complexity. Experiments across multiple VLM families including Qwen3-VL demonstrate consistent improvements. Our contributions are threefold:
\begin{itemize}
    \item A structured pairwise experience schema capturing discriminators, decision rules and reasoning error patterns---organized around diagnosis pairs to directly support differential reasoning.
    \item A two-phase memory construction process that extracts and refines experience from diagnostic errors, enabling continuous and robust accumulation without parameter updates.
    \item Comprehensive evaluation on an education-oriented radiology benchmark demonstrating consistent improvements, with analytical experiments validating experience quality and robustness.
\end{itemize}

\section{MedExpMem Framework}
\label{sec:method}

\subsection{Problem Formulation}
\label{subsec:formulation}

Given a patient's medical images $I$ and clinical history $x$, a VLM-based diagnostic agent $\mathcal{A}$ renders a diagnosis $d = \mathcal{A}(x, I)$. This formulation is static, where agent knowledge remains unchanged regardless of cases encountered. We extend this to a dynamic adapting paradigm:
\begin{equation}
    d = \mathcal{A}(x, I; \mathcal{M}), \quad e = \textsc{Extract}(x, I, d, d^*), \quad \mathcal{M} \leftarrow \mathcal{M} \cup \{e\}
    \label{eq:dynamic}
\end{equation}
where the agent first produces diagnosis $d$ conditioned on experience memory $\mathcal{M}$, then compares $d$ against expert ground truth $d^*$ to extract and accumulate experience memory $e$. This formulation enables continuous experience accumulation without parameter updates---particularly valuable for locally deployed models where fine-tuning is impractical due to privacy and computational constraints.

\begin{figure*}[t]
\centering
\includegraphics[width=0.98\textwidth]{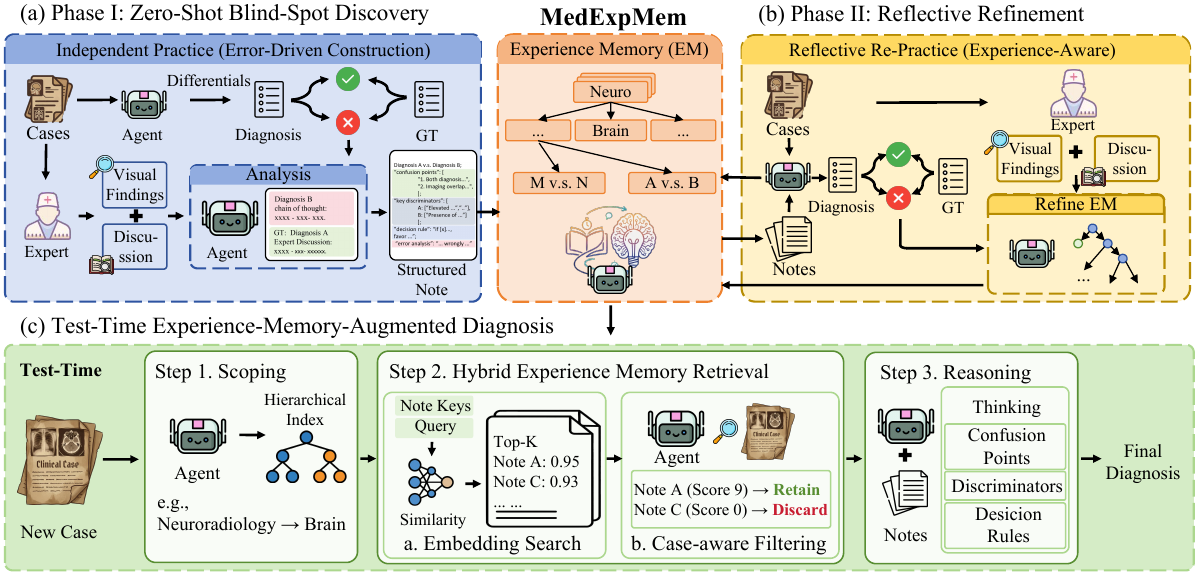}
\caption{Overview of the MedExpMem framework. \textbf{(a) Phase I: Zero-Shot Blind-Spot Discovery.} The agent conducts zero-shot diagnosis. \textbf{(b) Phase II: Reflective Refinement.} The agent re-diagnoses cases with experience memory access. \textbf{(c) Test-Time Inference.} Agent performs experience-memory-augmented reasoning with hybrid-retrieval.}
\label{fig:framework}
\vspace{-0.5cm}
\end{figure*}

\subsection{Pairwise Differential Experience Notes}
\label{subsec:note_structure}

We design structured experience notes organized around \textit{diagnosis pairs}, 
the fundamental unit of differential reasoning, rather than isolated diseases. This pairwise organization offers critical advantages over single-disease retrieval. First, it directly addresses differential diagnosis: when distinguishing confusable conditions, an ``A vs.\ B'' note provides targeted discriminative knowledge unavailable from separate descriptions of A and B. Second, it reflects the clinical reasoning pattern ``Is this A or B'' rather than ``What is A''~\cite{schmidt1990cognitive}. Third, it supports bidirectional reasoning: each note encodes features favoring \textit{both} diagnoses, preventing confirmation bias that single-disease retrieval may introduce.

Table~\ref{tab:note_structure} summarizes the note schema for experience memory. Each note is indexed by \texttt{differential\_pair}, e.g., ``Lymphoma vs.\ Metastasis'', representing the minimal discriminative unit in differential diagnosis. Core fields include \texttt{discriminators} capturing distinguishing features, \texttt{decision\_rule} providing actionable conditional guidance, and \texttt{error\_analysis} recording agent's own reasoning error patterns that led to past misdiagnosis. Unlike static knowledge bases describing diseases in isolation, these fields are derived from agent-specific failures, yielding experience tailored to the model's own blind spots. Decision rules adopt probabilistic formulations to acknowledge diagnostic uncertainty, and are directly applicable as conditional checks during agent reasoning.

\begin{table}[t]
\centering
\caption{Experience note schema for differential diagnosis.}
\label{tab:note_structure}
\small
\begin{tabular}{@{}p{2.4cm}p{9.4cm}@{}}
\toprule
\textbf{Field} & \textbf{Description [Format]} \\
\midrule
\multicolumn{2}{@{}l}{\textit{Metadata}} \\
\texttt{department} & Radiology subspecialty for hierarchical indexing [single value] \\
\texttt{organ/region} & Anatomical localization within department [single value] \\
\texttt{differentials} & Two diagnoses being compared, serving as retrieval key [``A vs.\ B''] \\
\midrule
\multicolumn{2}{@{}l}{\textit{Core Differential Knowledge}} \\
\texttt{confusions} & Reasons why these diagnoses are commonly confused---overlapping imaging features and clinical presentations [enumerated list] \\
\texttt{discriminators} & Imaging, clinical, and laboratory findings distinguishing each diagnosis [structured text with disease-specific sections] \\
\texttt{decision\_rule} & Actionable conditional rules [``If X$_1$ $\rightarrow$ favor A; If X$_2$ $\rightarrow$ exclude A; If Y$_1$ $\rightarrow$ favor B; If Y$_2$ $\rightarrow$ exclude B; If Z $\rightarrow$ consider others''] \\
\midrule
\multicolumn{2}{@{}l}{\textit{Self Reasoning Error Pattern}} \\
\texttt{error\_analysis} & Case-specific reasoning failure decomposition \\
\bottomrule
\end{tabular}
\end{table}

The experience memory notes are organized following standard radiology subspecialty divisions~\cite{abr_subspecialty}: the first level corresponds to \textbf{departments} (e.g., neuroradiology, thoracic), and the second level specifies \textbf{organs/regions} within each department. This mirrors clinical radiology structure and enables anatomically-scoped retrieval. Our taxonomy covers 11 departments and 118 organ/region categories, with \texttt{others} providing coverage for atypical cases.


\subsection{Two-Phase Experience Memory Construction}
\label{subsec:construction}

We construct experience memory through a two-phase process: initial practice exposing knowledge gaps, followed by reflective refinement.

\textbf{Phase I: Zero-shot Error Discovery.}
The agent first diagnoses educational cases without memory access, exposing intrinsic reasoning limitations. For each error where diagnosis $d \neq d^*$, we extract an experience note:
\begin{equation}
    e = \textsc{Extract}(x, I, d, d^*, \text{discussion})
\end{equation}
where \texttt{discussion} contains expert imaging findings and diagnostic reasoning. The agent is instructed to: (1) identify generalizable confusion points between $d$ and $d^*$, (2) extract discriminating features, (3) formulate bidirectional decision rules, and (4) analyze its reasoning failure. Memory inaccessibility during Phase I helps ensure notes reflect genuine knowledge gaps.

\textbf{Phase II: Reflective Refinement.}
The second phase re-diagnoses all cases with memory access, serving complementary purposes: (1) \textit{capturing stochastic errors}, cases correctly diagnosed by chance in may fail under repeated trials; (2) \textit{identifying note deficiencies}, when relevant notes are retrieved but errors persist, Phase II extraction supplements missing information; (3) \textit{detecting misleading notes}, if notes cause correct-to-incorrect transitions, error analysis enables refinement. New insights are merged with existing ones if differential pairs exist. 

\subsection{Experience-Memory-Augmented Diagnosis}
\label{subsec:diagnosis}

Inferencing on new cases, the agent performs diagnosis through three steps:

\textit{Step 1: Anatomical Scoping.}
Clinical cases involve multiple organ systems or cross departmental boundaries. The agent recalls experience memory's hierarchical index and selects 1--2 department-organ paths based on case presentation.

\textit{Step 2: Hybrid Experience Retrieval.}
Within selected paths, embedding similarities between candidate differential diagnoses and each note's \texttt{differential pair} keys are computed using a domain-specific encoder. Pairwise matching ensures symmetric relevance to both competing hypotheses, mitigating retrieval bias toward a single diagnosis. Notes exceeding similarity threshold $\tau$ are retrieved, and top-$K$ truncation is applied. The agent then scores each retrieved note on relevance, excluding clearly irrelevant notes (wrong organ system, unrelated disease category) while retaining potentially useful ones---a conservative strategy robust to smaller model limitations.

\textit{Step 3: Experience-Memory-Augmented Reasoning.}
Retained experience memories are integrated into the diagnostic prompt. The agent autonomously leverages notes to inform its differential reasoning and produce the final diagnosis.

\section{Experiments}
\label{sec:experiments}

\subsection{Experimental Setup}

\textbf{Dataset.} We construct our benchmark from Eurorad~\cite{eurorad}, an open-source peer-reviewed radiology teaching file maintained by the European Society of Radiology. Eurorad comprises over 10,000 cases spanning neuroradiology, thoracic, abdominal, musculoskeletal, cardiac, and interventional radiology. Each case provides structured \textit{clinical history}, \textit{multi-modality imaging} (CT, MRI, X-ray, ultrasound), \textit{expert-annotated imaging findings}, \textit{reference-based discussion}, \textit{ground-truth diagnosis}, and \textit{curated differential diagnoses} representing clinically plausible alternatives. Compared to classical medical VQA benchmarks~\cite{pathvqa,vqarad,slake,pmcvqa}, this education-oriented data with broader coverage is well-suited for targeted clinical experience accumulation. We collect 6,228 high-quality cases published before year 2025 for the \textit{construction phase}, during which agents build personalized experience libraries. For evaluation, we curate 227 high-quality cases from year 2025 as test set. This temporal split mirrors realistic clinical training: accumulating experience from historical cases before encountering new challenges.

\textbf{Implementation Details.} We employ S-PubMedBert-MS-MARCO~\cite{deka2022improved}, a domain-specific embedding model built upon PubMedBERT~\cite{gu2021domain}, for keyword embedding. We set similarity threshold $\tau=0.9$ and retrieve top-$K=10$ notes. All VLMs operate with default temperature settings. During construction, each agent independently builds its own experience memory, ensuring notes align with agent-specific knowledge gaps.

\subsection{Main Results}

Medical deployment demands privacy protection and local inference, favoring open-source models. We evaluate Qwen3-VL~\cite{qwen3vl}, InternVL3.5~\cite{internvl3_5}, and Lingshu~\cite{lingshu}, a medical-specialized VLM, across multiple scales.

Table~\ref{tab:main_results} reports diagnostic accuracy with and without experience memory. MedExpMem consistently improves performance across all models, demonstrating that structured experience complements parametric knowledge regardless of architecture or scale. Even the strongest baseline gains +4.7\%, showing experience addresses limitations beyond pre-training. Notably, the medical-specialized Lingshu-7B achieves lower baseline accuracy than subsequently released general-purpose models of comparable scale (Qwen3-VL-8B: 69.6\%). This observation highlights a limitation of static domain-specific fine-tuning: models trained on historical medical data may underperform on challenging cases that emerge beyond their knowledge cutoff.

\begin{table}[t]
\centering
\caption{Main results. \textbf{Baseline}: zero-shot diagnosis. \textbf{w/Exp}: diagnosis augmented with experience memory. 
\textbf{Recall}: proportion of cases with successfully retrieved notes. 
\textbf{Precision}: accuracy on cases where notes were retrieved. 
\textbf{Beneficial}: proportion of cases always corrected by experience memory (wrong $\rightarrow$ correct). 
\textbf{Harmful}: proportion of cases always misled by experience memory (correct $\rightarrow$ wrong).}
\label{tab:main_results}
\resizebox{\linewidth}{!}{
\begin{tabular}{ll ccc rccc}
\toprule
\multirow{2}{*}{\textbf{Domain}} & \multirow{2}{*}{\textbf{Model}} & \multicolumn{3}{c}{\textbf{Diagnostic Accuracy}} & \multicolumn{4}{c}{\textbf{Experience Memory}} \\
\cmidrule(lr){3-5} \cmidrule(lr){6-9}
 & & \textbf{Baseline} & \textbf{w/Exp} & \textbf{$\Delta$} & \textbf{Recall} & \textbf{Precision} & \textbf{Beneficial} & \textbf{Harmful}\\
\midrule
\multirow{5}{*}{General} 
 & Qwen3-VL-2B      & 54.6\% & 61.6\% & \textbf{+7.0\%} & 49.6\% & 60.7\% & 8.4\% & 1.3\% \\
 & Qwen3-VL-8B      & 69.6\% & 74.5\% & \textbf{+4.9\%} & 77.5\% & 74.4\% & 9.7\% & 4.8\% \\
 & Qwen3-VL-30B     & 74.9\% & 79.6\% & \textbf{+4.7\%} & 58.6\% & 80.5\% & 7.0\% & 2.2\% \\
\cmidrule{2-9}
 & InternVL3.5-8B   & 64.0\% & 71.0\% & \textbf{+6.9\%} & 81.9\% & 73.7\% & 12.7\% & 5.8\% \\
 & InternVL3.5-30B  & 74.0\% & 77.5\% & \textbf{+3.5\%} & 68.3\% & 75.5\% & 8.8\% & 5.2\% \\
\midrule
\multirow{2}{*}{Medical} 
 & Lingshu-7B       & 66.1\% & 67.8\% & \textbf{+1.7\%} & 58.2\% & 68.2\% & 3.5\% & 1.7\% \\
 & Lingshu-32B      & 72.2\% & 74.8\% & \textbf{+2.6\%} & 54.1\% & 74.2\% & 6.5\% & 3.8\%\\
\bottomrule
\end{tabular}
}
\end{table}

\subsection{Analytical Experiments}

\begin{table}[t]
\centering
\caption{Ablation study on Qwen3-VL-8B. Impact of construction rounds, retrieval scope, and similarity threshold. }
\label{tab:ablation}
\begin{tabular}{lcc}
\toprule
\textbf{Configuration} & \textbf{Accuracy} & \textbf{$\Delta$} \\
\midrule
Baseline (zero-shot) & 69.6\% & -- \\
w/ PubMed-RAG & 72.4\% & +2.8\% \\
w/ \textbf{MedExpMem} & \textbf{74.5\%} & \textbf{+4.9\%} \\
\midrule
\multicolumn{2}{@{}l}{MedExpMem Ablation} \\
One-round construction & 73.1\% & +3.5\% \\
Single-department retrieval & 72.2\% & +2.6\% \\
Threshold $\tau=0.80$ & 73.1\% & +3.5\% \\
Threshold $\tau=0.95$ & 72.8\% & +3.2\% \\
\bottomrule
\end{tabular}
\end{table}

We conduct ablation experiments on Qwen3-VL-8B to verify key design choices, and the results are reported in Table~\ref{tab:ablation}. \textbf{\textit{Two-round construction}} further improves performance by expanding and refining notes. \textbf{\textit{Cross-department search}} retrieval yields larger gains than single-department retrieval, reflecting realistic diagnostic complexity. For \textbf{\textit{retrieval threshold}}, performance peaks at $\tau=0.9$, validating the similarity threshold design.

Furthermore, we compare the performance with RAG technology. Specifically, we implement retrieval from PubMed~\cite{pubmed} and allow the agent to retrieve the top-3 related academic literature articles. However, the results show that textbook-like references provide limited gains to the agent when dealing with high challenging cases, as they increase retrieval scope and inference cost. This supports our hypothesis that agents benefit from customized experience memory to address their reasoning blind spots, rather than static textbook knowledge.

The experimental results in Table~\ref{tab:main_results} verify the effectiveness of experience memory. We observe a recall--scale trade-off: larger models retrieve fewer notes due to fewer prior errors, whereas smaller models sometimes fail to identify optimal retrieval paths. Cases with retrieved notes are typically more challenging, yet experience memory elevates their accuracy toward baseline levels. Although memory can introduce both beneficial and harmful effects, the overall gains outweigh the risks, which largely depend on note quality and retrieval precision.

\begin{figure*}[t]
\centering
\includegraphics[width=0.98\textwidth]{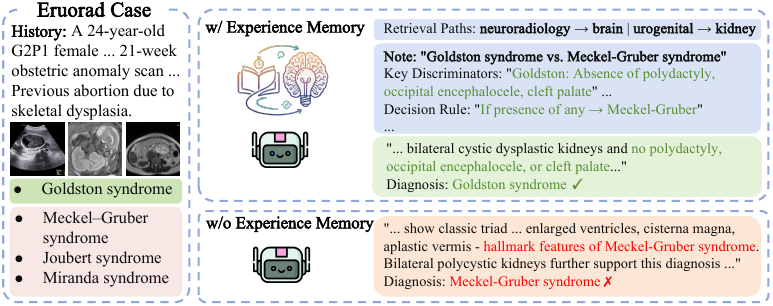}
\caption{Case study comparing diagnosis with and without experience memory. The retrieved pairwise note provides actionable discriminators that guide correct diagnosis.}
\label{fig:casestudy}
\vspace{-0.5cm}
\end{figure*}







\subsection{Case Study}

Figure~\ref{fig:casestudy} shows a fetal anomaly case with Dandy–Walker malformation and cystic kidneys, where the key differential lies between Goldston syndrome and Meckel–Gruber syndrome. With experience memory, the agent retrieved the pairwise note “Goldston vs.\ Meckel–Gruber”, which specifies that the absence of polydactyly, occipital encephalocele, and cleft palate favors Goldston syndrome. Recognizing these negative discriminators, the agent produced the correct diagnosis. Without memory, the agent focused on shared positive findings and predicted Meckel–Gruber syndrome. This case illustrates how pairwise experience memory guide absence-based differential reasoning, which can be difficult to achieve utilizing finetuning.


\section{Conclusion}

We introduced MedExpMem, an experience memory framework that enables vision-language diagnostic agents to accumulate structured differential expertise without parameter updates. By organizing knowledge as pairwise differential notes rather than isolated disease descriptions, MedExpMem shifts retrieval from static characterization to comparative clinical reasoning. A two-phase construction process identifies reasoning blind spots and refines them through reflective re-diagnosis, allowing experience to persist across cases. Experiments on a temporally split radiology benchmark demonstrate consistent improvements across model families and scales, demonstrating that structured experience addresses adaptation needs beyond the reach of parametric knowledge alone. 

\bibliographystyle{splncs04}

\end{document}